\documentclass[conference]{IEEEtran}
\IEEEoverridecommandlockouts
\usepackage{cite}
\usepackage{amsmath,amssymb,amsfonts}
\usepackage{algorithmic}
\usepackage{graphicx}
\usepackage{textcomp}
\usepackage{xcolor}

\def\BibTeX{{\rm B\kern-.05em{\sc i\kern-.025em b}\kern-.08em
    T\kern-.1667em\lower.7ex\hbox{E}\kern-.125emX}}

\usepackage[bookmarks=false]{hyperref}

\makeatletter

\def\ps@IEEEtitlepagestyle{
  \def\@oddfoot{\mycopyrightnotice}
  \def\@evenfoot{}
}
\def\mycopyrightnotice{
{\fbox{\parbox{\dimexpr\textwidth-\fboxsep-\fboxrule\relax}{
  {\textcopyright 2020 IEEE. Personal use of this material is permitted. Permission from IEEE must be obtained for all other uses, in any current or future media, including reprinting/republishing this material for advertising or promotional purposes, creating new collective works, for resale or redistribution to servers or lists, or reuse of any copyrighted component of this work in other works.}
  }}}
  \gdef\mycopyrightnotice{}
}

\usepackage{eso-pic}
\newcommand\AtPageUpperMyright[1]{\AtPageUpperLeft{
 \put(\LenToUnit{0.4\paperwidth},\LenToUnit{-1cm}){
     \parbox{0.6\textwidth}{\raggedleft\fontsize{9}{11}\selectfont #1}}
 }}
\newcommand{\conf}[1]{
\AddToShipoutPictureBG*{
\AtPageUpperMyright{#1}
}
}

\title{Multi-target regression via output space quantization\\
\thanks{This research is implemented through the Operational Program ``Human Resources Development, Education and Lifelong Learning'' and is co-financed by the European Union (European Social Fund) and Greek national funds.}
}
\conf{International Joint Conference on Neural Networks (IJCNN) 19-25 July, 2020}
    
\begin{document}

\author{
\IEEEauthorblockN{Eleftherios Spyromitros-Xioufis\textsuperscript{1,2}, Konstantinos Sechidis\textsuperscript{1,3} and Ioannis Vlahavas\textsuperscript{1}}
\IEEEauthorblockA{\textsuperscript{1}\textit{Department of Computer Science, Aristotle University of Thessaloniki, Greece} \\
\textsuperscript{2}\textit{Expedia, Geneva, Switzerland}\\
\textsuperscript{3}\textit{School of Computer Science, University of Manchester, UK}\\
{espyromi@csd.auth.gr, konstantinos.sechidis@manchester.ac.uk, vlahavas@csd.auth.gr}
}
}


\maketitle


\begin{abstract}
Multi-target regression is concerned with the prediction of multiple continuous target variables using a shared set of predictors.
Two key challenges in multi-target regression are: (a) modelling target dependencies and (b) scalability to large output spaces. 
In this paper, a new multi-target regression method is proposed that tries to jointly address these challenges via a novel problem transformation approach. 
The proposed method, called MRQ, is based on the idea of quantizing the output space in order to transform the multiple continuous targets into one or more discrete ones.
Learning on the transformed output space naturally enables modeling of target dependencies while the quantization strategy can be flexibly parameterized to control the trade-off between prediction accuracy and computational efficiency.
Experiments on a large collection of benchmark datasets show that MRQ is both highly scalable and also competitive with the state-of-the-art in terms of accuracy.
In particular, an ensemble version of MRQ obtains the best overall accuracy, while being an order of magnitude faster than the runner up method.
\end{abstract}


\section{Introduction}\label{sec:introduction}

Multi-target regression, also called multivariate or multi-output regression, is an instance of multi-target prediction \cite{waegeman2019multi} where the goal is to estimate multiple continuous variables based on a common set of predictors. Under the existence of statistical dependencies between the target variables, their joint modeling has been shown to be advantageous compared to modeling each variable independently \cite{hastie09}. This is for instance the case in applications such as stock prediction \cite{ghosn1997multi}, energy production forecasting in photovoltaic farms \cite{ceci2016predictive} and water quality monitoring \cite{hatzikos2008}.

Similarly to methods for multi-label classification \cite{tsoumakas2010b}, multi-target regression methods can be broadly categorized in two groups: (i) \textit{algorithm adaptation} and (ii) \textit{problem transformation}. Algorithm adaptation methods extend specific learning algorithms (e.g. support vector regression \cite{sanchez2004svm}, trees \cite{appice2007}, etc.) in order to handle multiple outputs, while problem transformation methods transform the learning task into one or more single output tasks that can be solved with existing learning algorithms. 
Algorithm adaptation methods often generate a single multi-output model that is easier to interpret and are more scalable to large output spaces compared to existing problem transformation methods. On the other hand, problem transformation methods can be easily adapted to the problem at hand by employing suitable base learners and have been found superior to algorithm adaptation methods in terms of accuracy \cite{spyromitros2016multi}.

Despite the fact that problem transformation methods have seen widespread use in multi-label classification (largely owing to their excellent predictive performance \cite{zhang2014}), their potential in the context of multi-target regression has only recently been explored. 
More specifically, methods that expand the input space of the independent regressions baseline using estimations of the other target variables as meta inputs were introduced in \cite{spyromitros2016multi}, while \cite{tsoumakas:2014:ecmlpkdd} proposed an ensemble approach that is based on constructing random linear combinations of the target variables. 
Both these methods were found competitive with the state-of-the-art but scale linearly (at best) with respect to the number of targets, thereby having limited applicability in problems with large output spaces.

In this paper we propose \textit{Multi-target Regression via Quantization} (MRQ), a novel problem transformation method that is based on the idea of using vector quantization \cite{gray1998quantization} in order to map the original real-valued output vectors into a finite set of prototype vectors or centroids. 
After this transformation, the multi-target regression problem is transformed into one of multi-class classification, where the task is to predict the centroid that lies closer to the actual output vector.
This type of reduction is motivated by the fact that in many real world problems, variables that are originally continuous in nature are discrete by observation and hence it is reasonable and convenient to model an appropriate discrete approximation \cite{chakraborty2015generating}.

Compared to existing problem transformation approaches, MRQ has the advantage that it directly models the joint distribution of the targets through a discrete approximation and has a complexity that is practically independent of the number of target variables, making it scalable to problems with very large output spaces.

In addition to the basic version of MRQ where a single quantizer is used to encode the whole output space, we also introduce an ensemble version of the method (eMRQ) where multiple quantizers are used, each one encoding a randomly selected subset of the target variables. 
Hence, in eMRQ the multi-target regression problem is transformed into multiple multi-class classification problems and the predictions are obtained by averaging multiple centroids. Although eMRQ has a higher computational complexity, it provides significantly higher accuracy as it strikes a better balance between the two main error components of the approach, i.e. the \textit{quantization} and the \textit{classification} error.

The rest of the paper is organized as follows. Section~\ref{sec:related} presents related work on multi-target regression and draws parallels between our approach and some widely used problem transformation approaches for multi-label classification.
Section~\ref{sec:method} describes MRQ and its ensemble version eMRQ, including a discussion of some interesting theoretical properties of the method.
Section~\ref{sec:experiments} describes the experimental setup and section~\ref{sec:results} presents and discusses the experimental results. 
Finally, section~\ref{sec:conclusion} concludes the paper and discusses directions for future work.

\section{Related Work}\label{sec:related}

\subsection{Algorithm adaptation methods}

The first multi-target regression methods were developed by statisticians in the 70s and the 80s with Reduced Rank Regression \cite{Izenman1975}, FICYREG \cite{van1980multivariate}, two-block PLS \cite{wold1985partial} and Curds and Whey \cite{breiman1997} being some of the most characteristic examples. As shown in \cite{breiman1997}, all these methods have the same generic form, under which the estimates obtained by applying ordinary least squares regression on the target variables, are modified by a shrinkage matrix (calculated differently in each method) in order to to provide a more accurate prediction, under the assumption that the targets are correlated.  

More recently, a number of multi-target methods were derived from the predictive clustering tree framework \cite{blockeel1998}. Predictive clustering trees differ from standard decision trees in that the variance and prototype functions are treated as parameters that can be instantiated to fit a variety of learning tasks, including multi-target regression. Following this approach, \cite{blockeel1999} developed muti-objective decision trees where the variance function is computed as the sum of the variances of the targets, and the prototype function is the vector mean of the target vectors of the training examples falling in each leaf. Other approaches that fall under the same framework are the ensembles of multi-objective decision trees proposed in \cite{kocev2007} and a number of rule learning algorithms whose primary focus is on improving model interpretability (e.g. \cite{aho2012}). 

Another large group of methods stem from a regularization perspective \cite{zhou_chen_ye_2012}\footnote{Most of these methods were originally developed to solve the more general learning task of multi-task learning \cite{caruana1997multitask} but are commonly applied to multi-target regression tasks as well.}. These methods minimize a penalized loss of the form $\displaystyle \min_W \mathcal{L}(W)+\Omega(W)$, where $\mathcal{L}(W)$ is an empirical loss calculated on the training data, $W$ is an estimated parameter matrix, and $\Omega(W)$ is a regularization term whose particular form depends on the underlying task relatedness assumption. 
Most methods assume that all tasks are related to each other \cite{argyriou2008convex,chen2010learning,obozinski2010joint}, while there are methods assuming that tasks are organized in structures such as clusters \cite{zhou2011clustered}, trees \cite{kim2010tree} and graphs \cite{chen2010graph}. 
A systematic analysis of the connections between these regularization-based methods and related techniques from the Gaussian Processes framework is provided in \cite{alvarez2011kernels}.

Finally, a number of methods for multi-target regression have been derived by extending artificial neural networks to handle multiple outputs (e.g. \cite{caruana1995learning,baxter1995learning,ghosn1997multi,collobert2008unified}).

As the aforementioned techniques do not involve an explicit transformation of the multi-target regression problem but rather extend specific learning paradigms to handle multi-target regression tasks directly, they are regarded as algorithm adaptation methods.

\subsection{Problem transformation methods}

Contrary to the majority of multi-target regression methods, MRQ follows a problem transformation approach. This type of approaches reduce the multi-target prediction problem into one or more single-target prediction problems, for which a multitude of well-developed algorithms are readily available. The challenge is then to devise effective reduction approaches. 

Problem transformation methods have been extensively studied in the context of multi-label classification with Binary Relevance (BR) and Label Powerset (LP) being two of the simplest but also  widely used and theoretically justified (see, e.g., \cite{dembczynski2012}) methods. BR transforms the multi-label classification task into $m$ binary classification tasks, one for each label, while LP reduces multi-label classification into multi-class classification by treating each label combination as a distinct class value. As discussed in \cite{dembczynski2012}, although BR ignores label dependencies, it is well tailored for losses whose risk minimizer can be expressed in terms of marginal distributions $P(Y_i | \mathbf{x}) (i=1,\ldots,m)$ such as Hamming loss. LP on the other hand, can be seen as a method to estimate the conditional joint distribution $P(\mathbf{Y}|\mathbf{X})$, and while its basic form is tailored for the subset 0-1 loss, it can be extended to any loss function.  

While BR and LP are reasonable baselines, they have been extended and outperformed by more recent approaches. Two notable examples that have achieved state-of-the-art performance and received significant attention in the multi-label classification literature are Classifier Chains (CC) \cite{read2011} and RAkEL \cite{tsoumakas2011a}. CC is an extension of BR which manages to model label dependencies by augmenting the input space of each binary classifier with extra features that correspond to other labels. 
RAkEL extends LP by building multiple LP classifiers on random subsets of the original label space and combining their predictions with voting. This way, RAkEL tackles a number of LP's limitations such as modeling of scarce label sets and increased computational cost in problems with many distinct label sets.

Recently, both CC and RAkEL have offered inspiration for developing analogous approaches in multi-target regression. 
Regressor Chains (RC) \cite{spyromitros2016multi} is a direct adaptation of CC where regressors are used as base models instead of binary classifiers. Equipped with a mechanism to deal with noise propagation at inference time, RC was shown to achieve state-of-the-art performance in multi-target regression. 
RAkEL, on the other hand, inspired RLC \cite{tsoumakas:2014:ecmlpkdd}, an approach that builds an ensemble of regression models, each one concerning a random linear combination of a random subset of the target variables. RLC uses a sparse random matrix to project the original output space into a new output space of higher dimensionality, where each transformed variable represents the linear combination of two or more of the original targets. At inference time, an overdetermined system of linear equations is solved to recover the original space. While RLC was found competitive with other multi-target regression approaches in terms of performance, it suffers increased computational complexity as it requires a number of regression models that is significantly larger than the number of targets to perform competitively.
Concurrently to RLC, an approach based on random output space projections was developed in \cite{joly2014random} in the context of multi-label classification. Differently from RLC which aimed at improving prediction accuracy, the main goal of that approach was to reduce learning time complexity while maintaining the accuracy of predictions.  

The proposed approach bares a number of striking parallels with LP and RAkEL. Firstly, each cluster centroid in MRQ can be considered as the equivalent of a labelset (or label combination) in multi-label clasification. In that sense, MRQ resembles LP as it uses a multi-class classifier to predict the most likely centroid. In fact, it is easy to show that MRQ becomes equivalent to LP when the quantization error becomes zero. 
Given that eMRQ extends MRQ in the same way that RAkEL extends LP, it is straightforward to see their resemblance. 
Contrarily to other problem transformation methods for multi-target regression which have a linear (RC) or higher (RLC) complexity with respect to the number of targets, MRQ and eMRQ have a practically constant complexity with respect to the number of targets and can be tuned to provide a good trade-off between prediction accuracy and computational efficiency by appropriate parameterization of the quantization scheme.

Recently, we developed a transformation approach for feature selection on multi-target data based on the idea of output space quantization and found it to perform competitively against other feature selection methods \cite{sechidis2019information}. To the best of our knowledge, this is the first time that this idea is applied in the context of multi-target regression. 

\section{Method}\label{sec:method}

\subsection{Background and notation}

\subsubsection{Multi-target regression}

Given a set of training examples $D_{train}=\{{\bf x}^n, {\bf y}^n \}_{n=1}^{N}$, where ${\bf x} = [x_1 \hdots x_d]$ and ${\bf y} = [y_1 \hdots y_m]$ are realizations of the joint random variables ${\bf X} = X_1 \hdots X_d$ and ${\bf Y} = Y_1 \hdots Y_m$, the goal in multi-target prediction is to induce a model $\bf{h}:\bf{X} \rightarrow \bf{Y}$ that given an input vector $\bf{x}$, predicts an output vector $\bf{\hat{y}} = \bf{h}(\bf{x})$ that closely approximates the true output vector $\bf{y}$. In multi-target regression, all the output variables $Y_j$ are continuous (i.e. ${\bf Y} \in R^m$) while the input variables $X_i$ can have a real, ordinal or nominal domain.

The baseline Single-Target (ST) approach consists of building an independent regression model $h_j: {\bf X} \rightarrow Y_j$ for each target variable. Despite the obvious limitation of ignoring dependencies between targets, when coupled with a strong base learner, ST is very competitive in both multi-label classification (called BR in this context) \cite{luaces2012binary} and multi-target regression \cite{spyromitros2016multi}, especially on target-wise decomposable loss functions \cite{dembczynski2012}.

\subsubsection{Vector quantization}

Vector quantization is a technique that has its roots in information theory and was originally used for analog-to-digital conversion and data compression \cite{gray1998quantization}. In vector quantization, the goal is to reduce the cardinality of the representation space of high-dimensional, real-valued input data, while minimizing an objective distortion criterion. Formally, a vector quantizer is a function $q: R^d \rightarrow \mathcal{C}$ that maps each $d$-dimensional vector ${\bf x} \in R^d$ to a vector $q({\bf x}) \in \mathcal{C}$ where $\mathcal{C}=\{{\bf c}_i\}_{i=1}^k$ is a finite set of reproduction values or centroids ${\bf c}_i \in R^d$. Typically, a vector quantizer seeks to minimize the squared error between the input vector ${\bf x}$ and its reproduction value $q({\bf x})$ and is learned using Lloyd's algorithm (k-means). 
In this work, VQ is applied to transform the output space in multi-target regression problems by replacing each output vector by a value that corresponds to the index of the quantizer centroid that lies closer to that vector (section~\ref{sec:mrq}). This way, multi-target regression is reduced to multi-class classification.

As the dimensionality of the vectors that we want to quantize increases, so does the number of centroids $k$ that are required to maintain a small quantization error. As $k$ increases, it becomes impossible to learn a quantizer using k-means due to the fact that both the learning complexity as well as the number of required training samples are several times $k$. To address this issue, more efficient quantization techniques such as Product Quantization (PQ) \cite{jegou2011product} are used when dealing with high-dimensional vectors. In PQ, the vectors are split into $s$ non-overlapping subvectors of dimensionality $d'=d/s$ and a distinct lower-complexity subquantizer $\mathcal{C}_j$ is learned on each subspace using k-means. The reproduction values of such a quantizer are defined as the concatenation of the centroids of the $s$ subquantizers, thus the product quantizer maps each original vector to a vector from the Cartesian product $\mathcal{C} = \mathcal{C}_1 \times \ldots \times \mathcal{C}_s$. Assuming that all subquantizers have the same number of reproduction values $k'$, the product quantizer effectively generates a quantizer with $(k')^s$ reproduction values. In this work, we use a quantization approach that is similar to PQ in order to improve the accuracy of our method in problems with large output spaces (section~\ref{sec:emrq}). 

\subsection{MRQ}\label{sec:mrq}

The main idea behind MRQ is the use of vector quantization in order to transform the multi-dimensional continuous output space $\mathbf{Y} = Y_1 \ldots Y_m, Y_j \in R$ into a uni-dimensional discrete output space $Z \in \{1,\ldots,k\}$, where the levels of the categorical variable $Z$ correspond to the indices of the centroids $\{{\bf c}_i\}_{i=1}^k$ of a vector quantizer $q(\cdot)$ learned using k-means on the original\footnote{Output variables are actually standardized to ensure equal variances before applying k-means.} output space $\mathbf{Y}$.
At training time, a mapping $h_{MC}: \mathbf{X} \rightarrow Z $ is learned using a multi-class classifier.
At inference time, given an unknown instance $\mathbf{x}$, the multi-class classifier is first applied to get $\hat{z}=h_{MC}(\mathbf{x})$ and the corresponding centroid ${\bf c}_{\hat{z}}$ is returned as the final prediction.

The squared error of MRQ can be written as:
\begin{equation}
 SE(\mathbf{h}_{MRQ}) = \sum_{\{\mathbf{x},\mathbf{y}\} \in D_{test}}
\begin{cases}
 ||q(\mathbf{y})-\mathbf{y}||^2 & \text{when ${\hat{z}}={z}$} \\
 ||{\bf c}_{\hat{z}}-\mathbf{y}||^2 & \text{when ${\hat{z}}\neq{z}$} \\
\end{cases}
\end{equation}
Assuming that the quantizer $q$ satisfies the first Lloyd optimality condition, i.e that all vectors are quantized to their nearest centroids in terms of Euclidean distance: 
\begin{equation}
q(x) = \arg\min_{c_i \in C}||\mathbf{y}-\mathbf{c}_i||^2,
\end{equation}
it is easy to see that the lower bound for the squared error of MRQ is equal to the squared error of the underlying quantizer. Thus, using a quantizer with a small quantization error is a necessary condition for good performance in MRQ.
However, this lower bound is realized only when $h_{MC}$ is an oracle classifier that always predicts the correct class and any classification error can cause an arbitrarily large increase. Thus, achieving a small classification error $CE(h_{MC})= \sum_{\{\mathbf{x},z\} \in D_{test}} 1(\hat{z} \neq {z})$ is also crucial.

Interestingly, both error components of MRQ (i.e. the quantization and the classification error) are highly dependent on the parameter $k$. The quantization error, on one hand, is a monotonically decreasing function of $k$ for a Lloyd optimal quantizer. The classification error, on the other hand, is expected to increase with an increasing number of classes and a decreasing number of examples per class. Thus, finding a $k$ that strikes a good balance between the two error components is crucial for achieving good performance in MRQ.

\begin{figure}[h]
\centering
\resizebox{0.45\textwidth}{!}{
    \includegraphics[trim = 0.0cm 0.0cm 0.0cm 0.0cm, clip]{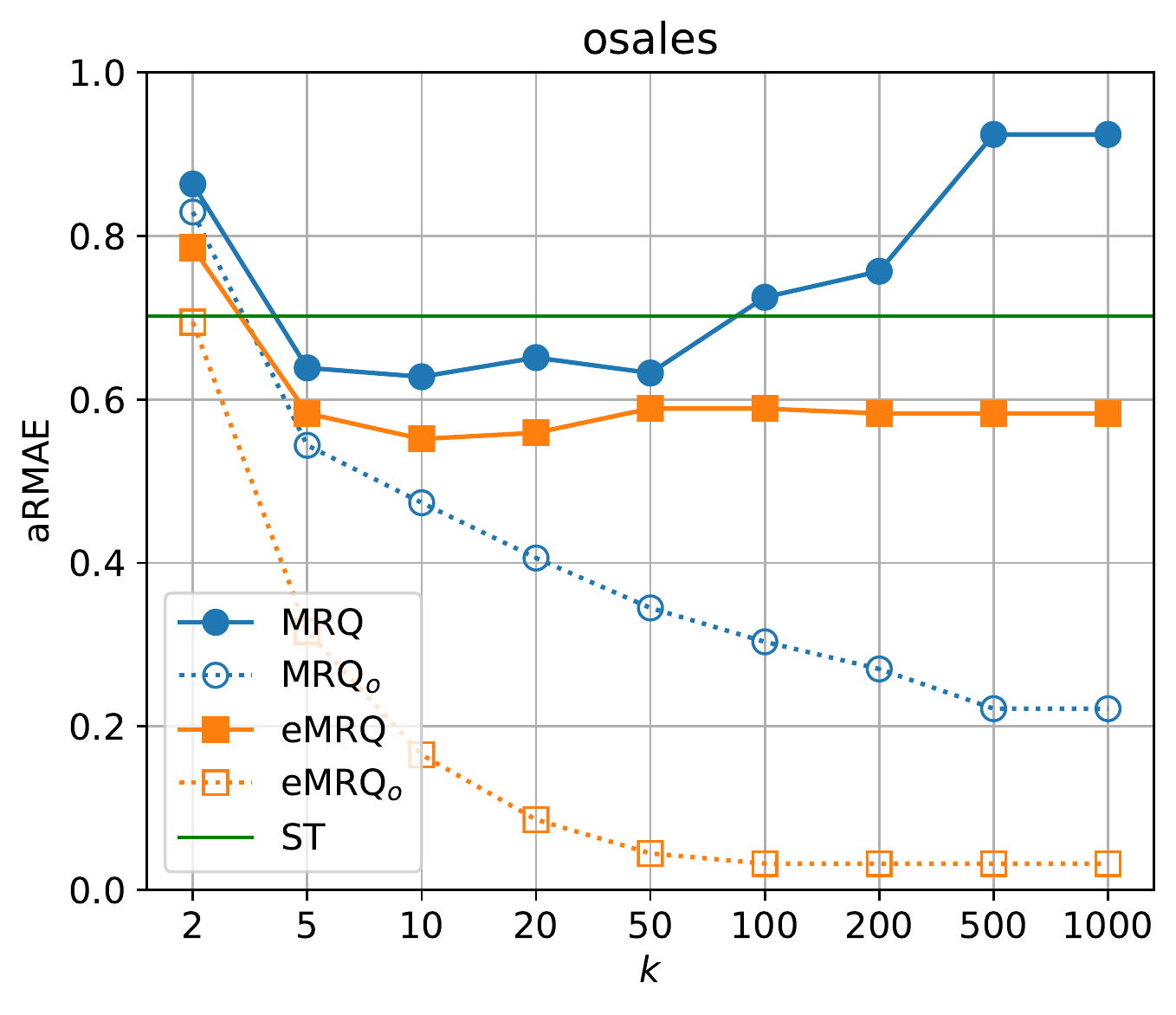}
}
\resizebox{0.45\textwidth}{!}{
    \includegraphics[trim = 0.0cm 0.0cm 0.0cm 0.0cm, clip]{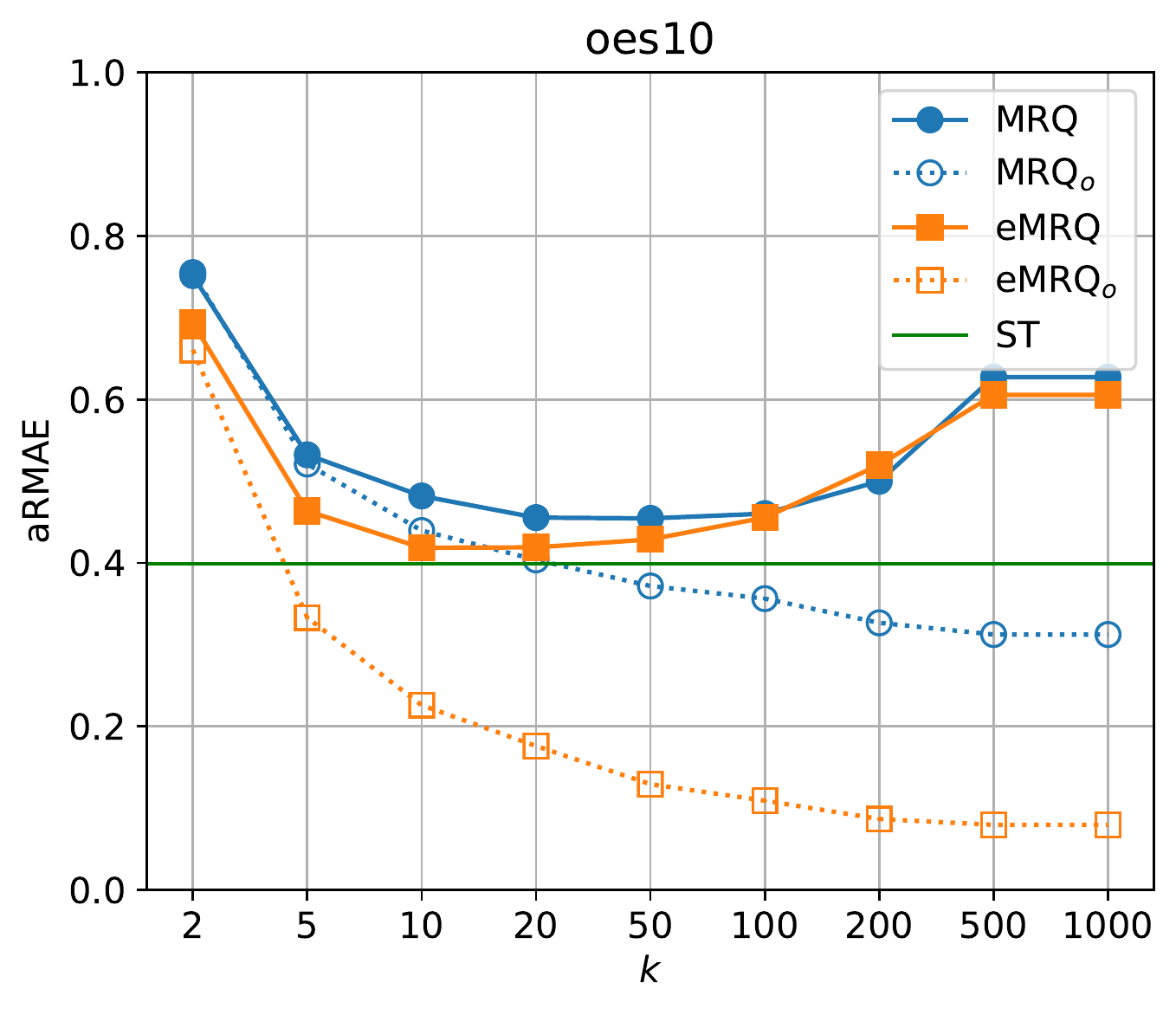}
}
\caption{Performance (aRMAE) of MRQ, MRQ$_o$ (oracle version of MRQ),  eMRQ, eMRQ$_o$ (oracle version of eMRQ) and ST as a function of $k$ on \texttt{osales} and \texttt{oes10}.}
\label{fig:mrq_oracle}
\end{figure}

This trade-off is highlighted in Figure~\ref{fig:mrq_oracle} which plots the error of MRQ (in two datasets), using an ensemble of classification trees as the multi-class classifier, against that of MRQ$_o$, an oracle version of the method whose underlying multi-class classifier is assumed to provide perfect predictions (i.e. $CE(h_{MC})=0$). We see that while the error of MRQ$_o$ decreases with $k$, the error of MRQ decreases initially as a result of a decreasing quantization error but then starts to increase as the classification error begins to dominate. We also notice that while in \texttt{osales} MRQ$_o$ has a smaller error than ST for $k >= 5$, $k >= 50$ is required in \texttt{oes10}. This is reflected in the performance of MRQ as in \texttt{osales} it obtains a smaller error than ST for a wide range of $k$ values, while in \texttt{oes10} it is always worse. 

\subsection{eMRQ}\label{sec:emrq}

eMRQ extends MRQ by employing a PQ-like approach to quantize the output space. More concretely, instead of building a single k-means quantizer $\mathcal{C}$ on the whole output space, eMRQ builds $s$ subquantizers $\{\mathcal{C}_j\}_{j=1}^{s}$ on variables ${\bf W}_j$ which are defined as random subsets of the original targets, i.e. ${\bf W}_j \subset {\bf Y} ~\forall~ j=1\dots s$. Contrarily to PQ where only disjoint subsets are considered, in eMRQ we allow the same target variable to participate in multiple groups and thus be redundantly quantized.

By employing this quantization scheme, eMRQ effectively transforms the mutli-target regression problem into $s$ multi-class classification subproblems, where the categorical class variables $Z_j \in \{1,\ldots,k\}$ encode the indices of the centroids of the subquantizers $\mathcal{C}_{j}$.
At inference time, all multi-class classifiers $h^{j}_{MC}: \mathbf{X} \rightarrow Z_j$ are queried to predict the corresponding targets ${\bf W}_j$ and the predictions for each of the original targets $Y_j$ are obtained by average pooling\footnote{For a predicate P, the expression [[P]] evaluates to 1 if P is true and to 0 if P is false.}:
$$
\hat{y}_j = 
\frac{\sum_{i=1}^{s} ([[Y_j \in \mathbf{W}_i]] * \hat{w}_{ij})}{\sum_{i=1}^{s} [[Y_j \in \mathbf{W}_i]]}
$$
where (with a slight abuse of notation) $\hat{w}_{ij}$ denotes the component of $\hat{w}_i$ that corresponds to variable $Y_j$.

eMRQ has three parameters: (a) the number of centroids $k$ in each subquantizer, (b) the number of original targets considered in each subquantizer, i.e. $|\mathbf{W}_j|$, henceforth denoted $NoT$ and (c) the total number of quantizers $s$. We observe that $NoT$ can take values in $\{1,\ldots,m\}$ and controls the degree of joint modelling of the target variables in the ensemble. The special case where $NoT=1$ corresponds to building an independent model for each target variable, while $NoT=m$ makes eMRQ equivalent to MRQ. We further observe that each of the original target variables is expected to participate in $\frac{s*NoT}{m}$ of the subproblems induced by eMRQ and therefore an $s \geq \frac{m}{NoT}$ should be used to guarantee that, on average, all target variables will be considered in at least one of the subproblems. 

Similarly to PQ, the quantization approach adopted by eMRQ allows it to implicitly induce a quantizer with $k^s$ reproduction values\footnote{Since the subquantizers in eMRQ are overlapping, the effective number of reproduction values will be smaller than $k^s$.}. Thus, it requires quantizers with a significantly smaller number of centroids than those of the single quantizer in MRQ to achieve the same quantization error. In turn, this has a positive impact on the classification error of eMRQ as each multi-class classification problem becomes simpler (fewer classes, more training examples per class). This effect can be seen in Figure~\ref{fig:mrq_oracle} where we notice that eMRQ$_o$, the oracle version of eMRQ (parameterized here with $NoT=2$ and $q=m$ and using the same multi-class classifier as MRQ), obtains a significantly smaller error than MRQ$_o$ for the same values of $k$. This results in eMRQ outperforming MRQ in both datasets.

\subsection{Computational Complexity}

The complexities of both MRQ and eMRQ depend on the complexities of the underlying multi-class classification and quantization algorithms. 
Given a multi-class classifier with training complexity $O(g(n,d,c))$ for a dataset with $n$ examples, $d$ input variables and $c$ classes, the complexity of MRQ is $O(t {\cdot} k {\cdot} n {\cdot} m) + O(g(n,d,k))$, where  $O(t {\cdot} k {\cdot} n {\cdot} m)$ is the complexity of running k-means for $t$ iterations. 
Similarly, the complexity of eMRQ is $O(s {\cdot} t {\cdot} k {\cdot} n {\cdot} NoT) + O(s {\cdot} g(n,d,k))$, since it learns $s$ quantizers on $NoT$-dimensional vectors and $s$ multi-class classifiers.
In practise, the training complexity of both methods is dominated by the complexity of building the multi-class classifier. As shown in section~\ref{sec:results}, by using a multi-class classifier with sublinear complexity with respect to the number of classes (i.e. an ensemble of decision trees), both methods are significantly faster than other problem transformation approaches.

\section{Experimental setup}\label{sec:experiments}

In this section we describe our experimental setup. We first present the datasets and their main characteristics, then provide the details of the setup used for MRQ, eMRQ and the other competing methods, and finally describe the evaluation methodology and the approach used to check for statistically significant performance differences. 

\subsection{Datasets}

The experiments are carried out on a large and diverse collection of multi-target regression datasets\footnote{\url{http://mulan.sourceforge.net/datasets-mtr.html}}, whose main characteristics are summarized in Table~\ref{tbl:data sets} (see \cite{spyromitros2016multi} for a detailed description of each dataset).
In addition to the number of examples, features and targets in each dataset, Table~\ref{tbl:data sets} also reports $\overline{|r|}$ and $\overline{H(\mathbf{Y})}$. 
$\overline{|r|}$ is the average Pearson correlation coefficient between all distinct pairs of targets in each dataset and is a rough measure of target interdependence.
$\overline{H(\mathbf{Y})}$ is the average entropy of the probability density functions (PDF) of the target variables, where each PDF is calculated by applying kernel density estimation using a gaussian kernel and a bandwidth tuned to maximize likelihood using 3-fold cross-validation.
$\overline{H(\mathbf{Y})}$ aims to serve as a measure of the expected quantization error in each dataset, as lower values point to a distribution that is concentrated around few specific values, while higher values point to a dispersed distribution (uniform being the most dispersed distribution having an entropy of 4.61 in this setup).

\begin{table}
\caption{Name, number of examples, number of input variables ($d$), number of target variables ($m$), average pairwise Pearson correlation ($\overline{|r|}$) and average output space entropy $\overline{H(\mathbf{Y})}$ of the datasets used in the evaluation.}
\label{tbl:data sets}
\begin{center}
\begin{tabular}{lrrrrr}
\hline
Name & \# ex. & $d$ & $m$ & $\overline{|r|}$ & $\overline{H(\mathbf{Y})}$\\ 
\hline
   edm &  154 &  16 &  2 & 0.01 & 3.06 \\
   enb &  768 &   8 &  2 & 0.98 & 4.51 \\
  jura &  359 &  15 &  3 & 0.20 & 3.97 \\
  scpf & 1137 &  23 &  3 & 0.73 & 2.35 \\
   sf1 &  323 &  10 &  3 & 0.23 & 1.66 \\
   sf2 & 1066 &  10 &  3 & 0.20 & 1.31 \\
 slump &  103 &   7 &  3 & 0.42 & 4.37 \\
 andro &   49 &  30 &  6 & 0.40 & 4.52 \\
 atp1d &  337 & 411 &  6 & 0.82 & 3.91 \\
 atp7d &  296 & 411 &  6 & 0.64 & 3.90 \\
   rf1 & 9125 &  64 &  8 & 0.39 & 3.96 \\
   rf2 & 9125 & 576 &  8 & 0.39 & 3.96 \\
osales &  639 & 401 & 12 & 0.62 & 2.44 \\
    wq & 1060 &  16 & 14 & 0.10 & 2.13 \\
 oes10 &  403 & 298 & 16 & 0.82 & 2.90 \\
 oes97 &  334 & 263 & 16 & 0.79 & 3.02 \\
 scm1d & 9803 & 280 & 16 & 0.64 & 4.07 \\
scm20d & 8966 &  61 & 16 & 0.60 & 4.03 \\
\hline
\end{tabular}
\end{center}
\end{table}

\subsection{Methods and parameters}
\label{sec:experiments:methods}

In section~\ref{sec:results}, the performance of MRQ and eMRQ is compared to the performance of ST, RLC \cite{tsoumakas:2014:ecmlpkdd} and ERC \cite{spyromitros2016multi}. Similarly to MRQ and eMRQ, ST, RLC and ERC take a problem transformation approach to multi-target regression and have been found significantly better than algorithm adaptation approaches such as ensembles of multi-objective decision trees (e.g. \cite{kocev2007}) and multi-task learning methods (e.g. \cite{argyriou2008convex}) in a previous empirical study \cite{spyromitros2016multi}. 

RLC and ERC are parametrized using the setup that leads to the best results according to the corresponding papers, i.e. 100 random linear combinations of 2 target variables are used in RLC, while in ERC we use the variant that generates out-of-sample estimates with 10 internal cross-validation folds and an ensemble size of 10. 
A crucial factor for the performance of any problem transformation approach is the underlying base learning algorithm. Throughout this study, ST, RLC and ERC are instantiated using a Bagging \cite{breiman1996} ensemble of 100 regression trees as the base regressor, following the recommendations of \cite{spyromitros2016multi}. To keep the comparison of MRQ and eMRQ with these methods as fair as possible, we instantiate them with a Bagging ensemble of 100 classification trees.

The proposed methods are implemented in Java and are integrated in the Mulan library \cite{tsoumakas2011b} which already contains implementations of ST, RLC and ERC. Thus, all methods are evaluated under a common framework\footnote{\url{https://github.com/lefman/mulan-extended}}.

\subsection{Evaluation methodology}
\label{sec:experiments:evaluation}

The performance of the methods is measured using average Relative Mean Absolute Error (aRMAE). The aRMAE of a model $\mathbf{h}$ on a dataset $D$ is defined as:
$$
aRMAE(\mathbf{h},D)= \frac{1}{m} \sum_{j=1}^{m}
\frac
{
    \sum_{(\mathbf{x},\mathbf{y}) \in D} |\hat{y}_{j}-y_{j}|
}
{
    \sum_{(\mathbf{x},\mathbf{y}) \in D} |\bar{Y}_{j}-y_{j}|
}
$$
where $\bar{Y}_{j}$ is the mean value of $Y_{j}$ over $D$ and $\hat{y}_{j}$ is the prediction of $\mathbf{h}$ for $Y_{j}$. Intuitively, aRMAE measures how much better (aRMAE\textless1) or worse (aRMAE\textgreater1) the model $\mathbf{h}$ is (on average) compared to a naive baseline that always predicts the mean value of each target.
To estimate aRMAE we use either repeated random subsampling (with 90\% of the data used for training and 10\% for validation) (sections~\ref{sec:results_mrq} and \ref{sec:results_emrq}) or $k$-fold cross-validation (section~\ref{sec:results:soa}). 

To test the statistical significance of the observed differences between the methods, we follow the methodology suggested by \cite{demsar2006} for comparing multiple methods on multiple datasets, i.e. we use the Friedman test to check the validity of the null-hypothesis (all methods are equivalent) and when the null-hypothesis is rejected ($p < 0.01$), we proceed with the Nemenyi post-hoc test. Instead of reporting the outcomes of all pairwise comparisons, we employ the simple graphical presentation of the test's results introduced in \cite{demsar2006}, i.e. all methods being compared are placed in a horizontal axis according to their average ranks and groups of methods that are not significantly different (at a certain significance level) are connected (see Figure~\ref{fig:soa_cd} for an example).
To generate such a diagram, a critical difference (CD) needs to be calculated that corresponds to the minimum difference in average ranks required for two methods to be considered significantly different. CD for a given number of methods and datasets, depends on the desired significance level. Due to the known conservancy of the Nemenyi test \cite{demsar2006}, we use a 0.05 significance level for computing the CD throughout the paper.

\section{Results and discussion}
\label{sec:results}

\subsection{Empirical evaluation of MRQ}
\label{sec:results_mrq}

In this section we study the performance of MRQ for different values of its parameter $k$ and compare it against ST.
The results are summarized in Table~\ref{tbl:mrq} which shows the error of MRQ for $k\in\{2, 5, 10, 20, 50, 100, 200, 500, 1000\}$ as well as the error of ST on each dataset. The best performance per dataset is highlighted in bold, while the best performance among the MRQ variants is underlined\footnote{Note that whenever $k$ is larger than the number of distinct output vectors $card$, $k=card$ is used.}. 

We see that the choice of $k$ can have a big impact on the method's performance, as different values lead to optimal results on each dataset.
Trying to shed some light into the factors that affect the optimal value of $k$, we observe that datasets with larger output spaces typically benefit from larger values of $k$ and vice versa. As a result, the best performance on \texttt{scm1d} and \texttt{scm20d} which have the highest number of targets in the collection is obtained with $k=1000$, while the best performance on \texttt{edm} which has the smallest number of targets is obtained with $k=5$. However, there are notable exceptions to this rule: (a) \texttt{enb} has the same number of targets as \texttt{edm} but the best performance is obtained with $k=200$, (b) \texttt{oes10} and \texttt{oes97} have the same number of targets as \texttt{scm1d} and \texttt{scm20d} but the best performance is obtained with $k=50$. In the case of \texttt{enb}, we observe that it is among the two datasets with the highest output space entropy. Hence, despite the small number of targets, a high $k$ is required to reduce the quantization error. In the case of \texttt{oes10} and \texttt{oes97}, on one hand these datasets have relatively small output space entropies and, on the other hand, they have a small number of examples (403 and 334 respectively) which makes it difficult to learn an accurate quantizer with $k>50$.
Summarizing the above observations, we conclude that the optimal value of $k$ is affected by the following factors: (a) the number of targets, (b) the entropy of the output space, (c) the number of training examples.

\begin{figure}
\centering
\resizebox{0.48\textwidth}{!}{
\includegraphics[trim = 1.2cm 0.7cm 1.2cm 0.3cm, clip]{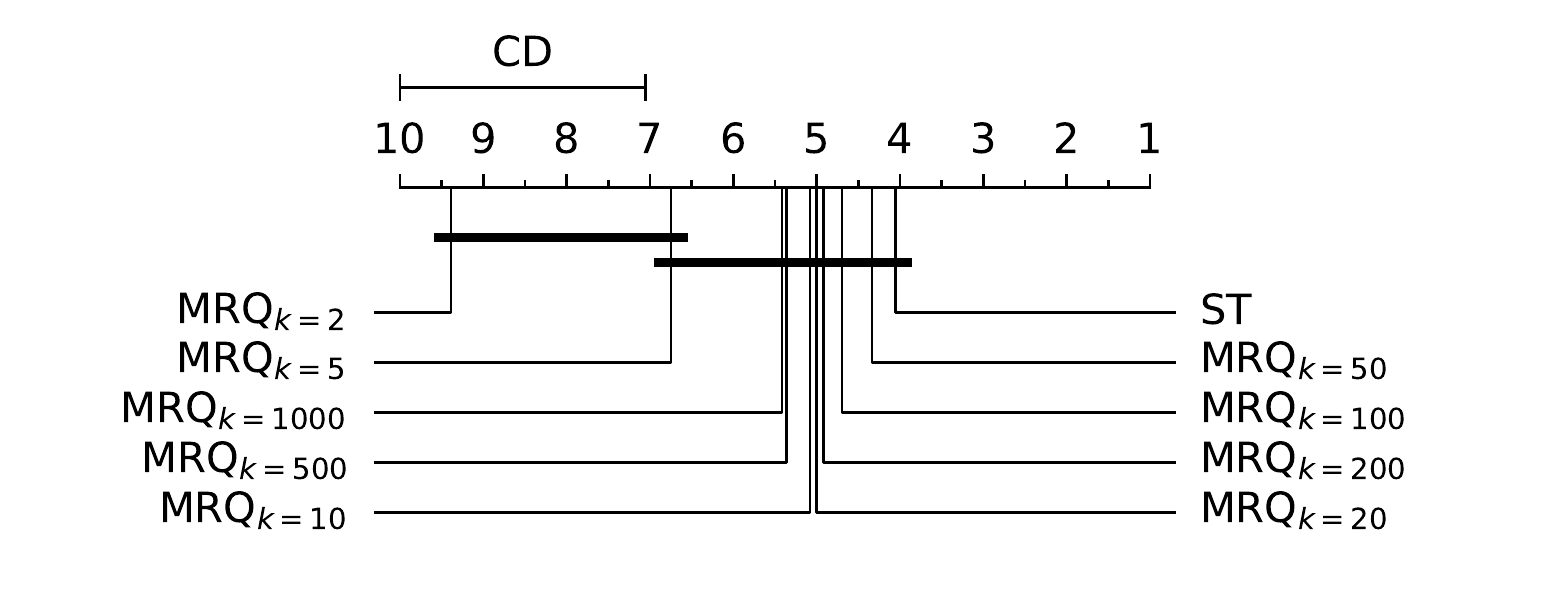}
}
\caption{Critical difference diagram between ST and MRQ variants.}
\label{fig:mrq_cd}
\end{figure}

Comparing the performance of MRQ to that of ST, we observe that in 11 out of 18 datasets ST is outperformed by one of the MRQ variants, which suggests that MRQ can be very competitive with appropriate parametrization. In fact, even with a fixed $k$, we see (last row of Table~\ref{tbl:mrq}) that MRQ with $k=50$ obtains a similar average rank with ST (4.333 vs 4.056). As shown in the critical difference diagram of Figure~\ref{fig:mrq_cd}, only the variant that uses $k=2$ is found statistically significantly worse than ST.

\setlength\tabcolsep{1.7pt} 

\begin{table}
\caption{MRQ performance for different values of $k$.  The best performance on each dataset is highlighted in bold and the best performance among MRQ variants is underlined. The last row shows the average rank of each method.}
\label{tbl:mrq}
\begin{center}
\begin{tabular}{l| rrrrrrrrr|r}
\hline
&  \multicolumn{9}{c|}{MRQ} \\
\cline{2-10}
Dataset & $k$=2 & 5 & 10 & 20 & 50 & 100 & 200 & 500 & 1000 & ST\\
\hline
   edm & 1.076 & \underline{\textbf{0.545}} & \underline{\textbf{0.545}} & \underline{\textbf{0.545}} & \underline{\textbf{0.545}} & \underline{\textbf{0.545}} & \underline{\textbf{0.545}} & \underline{\textbf{0.545}} & \underline{\textbf{0.545}} & 0.840 \\
   enb & 0.350 & 0.217 & 0.151 & 0.127 & 0.119 & 0.110 & \underline{0.102} & 0.115 & 0.177 & \underline{\textbf{0.085}} \\
  jura & 0.880 & 0.709 & 0.664 & 0.649 & 0.617 & \underline{0.610} & 0.639 & 1.008 & 1.008 & \underline{\textbf{0.511}} \\
  scpf & 0.835 & 0.516 & \underline{\textbf{0.473}} & 0.490 & 0.514 & 0.542 & 0.533 & 0.525 & 0.525 & 0.607 \\
   sf1 & 0.695 & 0.591 & \underline{\textbf{0.581}} & \underline{\textbf{0.581}} & \underline{\textbf{0.581}} & \underline{\textbf{0.581}} & \underline{\textbf{0.581}} & \underline{\textbf{0.581}} & \underline{\textbf{0.581}} & 0.957 \\
   sf2 & 0.590 & 0.454 & \underline{\textbf{0.408}} & \underline{\textbf{0.408}} & \underline{\textbf{0.408}} & \underline{\textbf{0.408}} & \underline{\textbf{0.408}} & \underline{\textbf{0.408}} & \underline{\textbf{0.408}} & 0.843 \\
 slump & 0.883 & 0.838 & 0.844 & \underline{0.821} & 0.839 & 0.950 & 0.950 & 0.950 & 0.950 & \underline{\textbf{0.703}} \\
 andro & 0.922 & 0.472 & \underline{\textbf{0.446}} & 0.546 & 0.685 & 0.685 & 0.685 & 0.685 & 0.685 & 0.476 \\
 atp1d & 0.614 & 0.481 & 0.442 & 0.402 & 0.360 & 0.307 & 0.318 & \underline{\textbf{0.285}} & \underline{\textbf{0.285}} & 0.325 \\
 atp7d & 0.805 & 0.575 & 0.489 & 0.430 & 0.394 & \underline{\textbf{0.287}} & 0.332 & 0.332 & 0.332 & 0.429 \\
   rf1 & 0.792 & 0.529 & 0.385 & 0.275 & 0.180 & 0.121 & 0.080 & 0.045 & \underline{\textbf{0.031}} & 0.045 \\
   rf2 & 0.792 & 0.529 & 0.385 & 0.275 & 0.180 & 0.121 & 0.080 & 0.045 & \underline{\textbf{0.031}} & 0.050 \\
osales & 0.864 & 0.639 & \underline{\textbf{0.628}} & 0.651 & 0.632 & 0.725 & 0.757 & 0.924 & 0.924 & 0.702 \\
    wq & 0.950 & 0.899 & 0.888 & 0.897 & \underline{0.884} & 0.885 & 0.890 & 0.918 & 0.961 & \underline{\textbf{0.858}} \\
 oes10 & 0.752 & 0.532 & 0.482 & \underline{0.455} & \underline{0.455} & 0.460 & 0.500 & 0.627 & 0.627 & \underline{\textbf{0.399}} \\
 oes97 & 0.843 & 0.668 & 0.637 & 0.635 & \underline{0.632} & 0.685 & 0.755 & 0.822 & 0.822 & \underline{\textbf{0.570}} \\
 scm1d & 0.704 & 0.566 & 0.512 & 0.458 & 0.389 & 0.345 & 0.307 & 0.266 & \underline{0.241} & \underline{\textbf{0.236}} \\
scm20d & 0.740 & 0.614 & 0.557 & 0.504 & 0.416 & 0.359 & 0.324 & 0.278 & \underline{\textbf{0.260}} & 0.334 \\
\hline
Av. rank & 9.389 & 6.750 & 5.083 & 5.000 & \underline{4.333} & 4.694 & 4.917 & 5.361 & 5.417 & \underline{\textbf{4.056}} \\
\hline
\end{tabular}
\end{center}
\end{table}

\subsection{Empirical evaluation of eMRQ}
\label{sec:results_emrq}

\begin{figure*}[h]
\centering
\resizebox{0.32\textwidth}{!}{
    \includegraphics[trim = 0.0cm 0.1cm 0.1cm 0.2cm, clip]{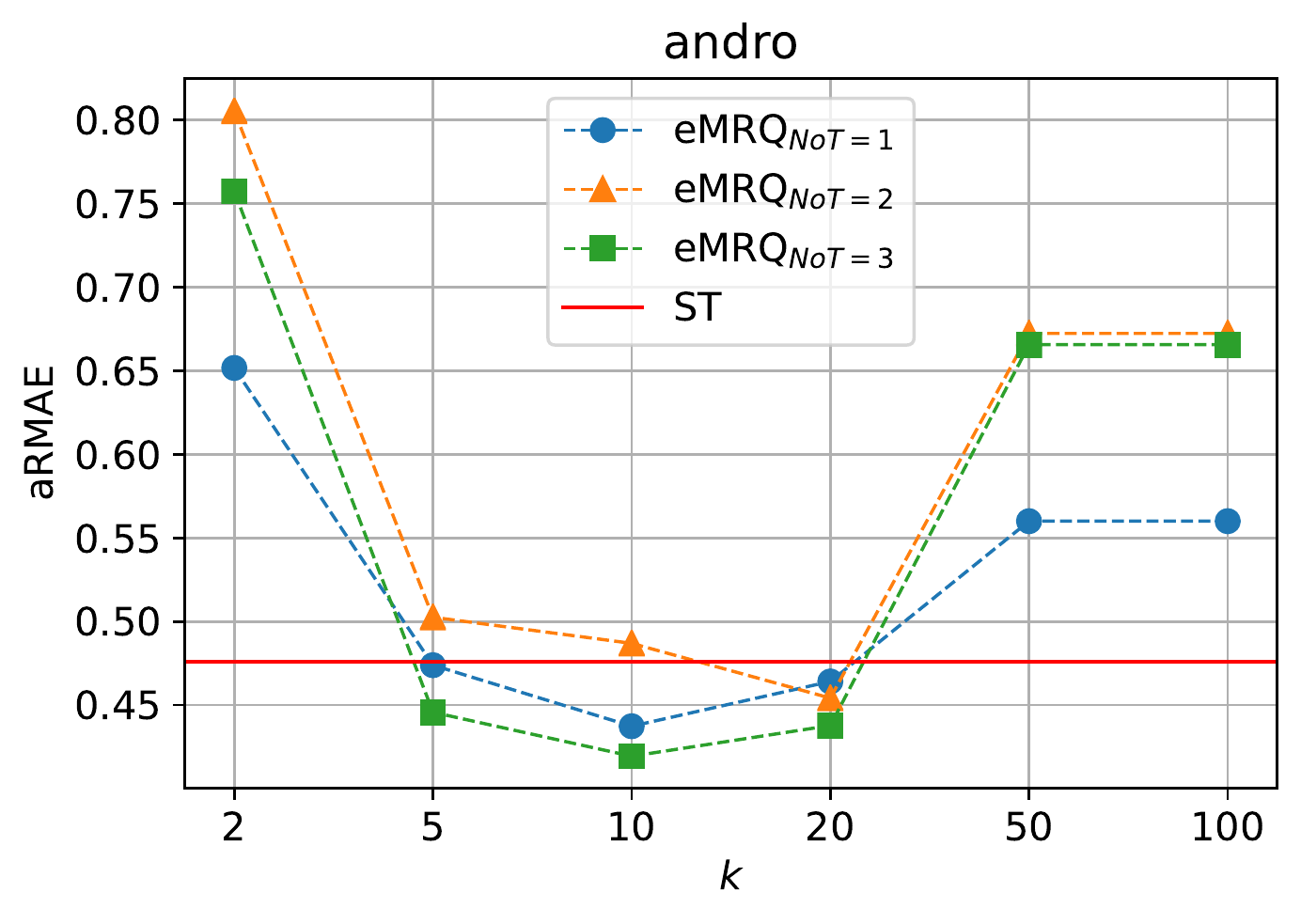}
}
\resizebox{0.32\textwidth}{!}{
    \includegraphics[trim = 0.0cm 0.1cm 0.1cm 0.2cm, clip]{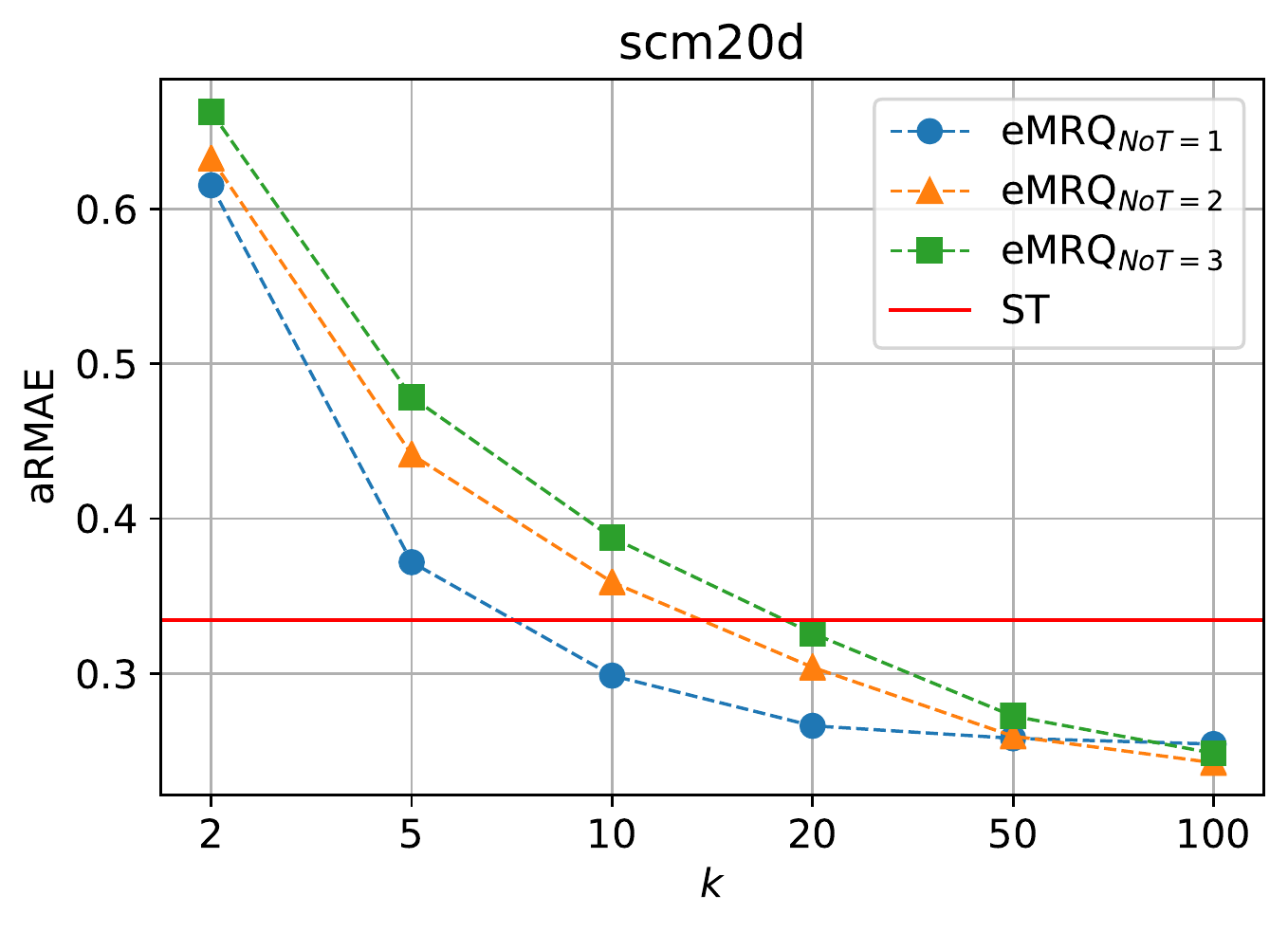}
}
\resizebox{0.32\textwidth}{!}{
    \includegraphics[trim = 0.0cm 0.1cm 0.1cm 0.2cm, clip]{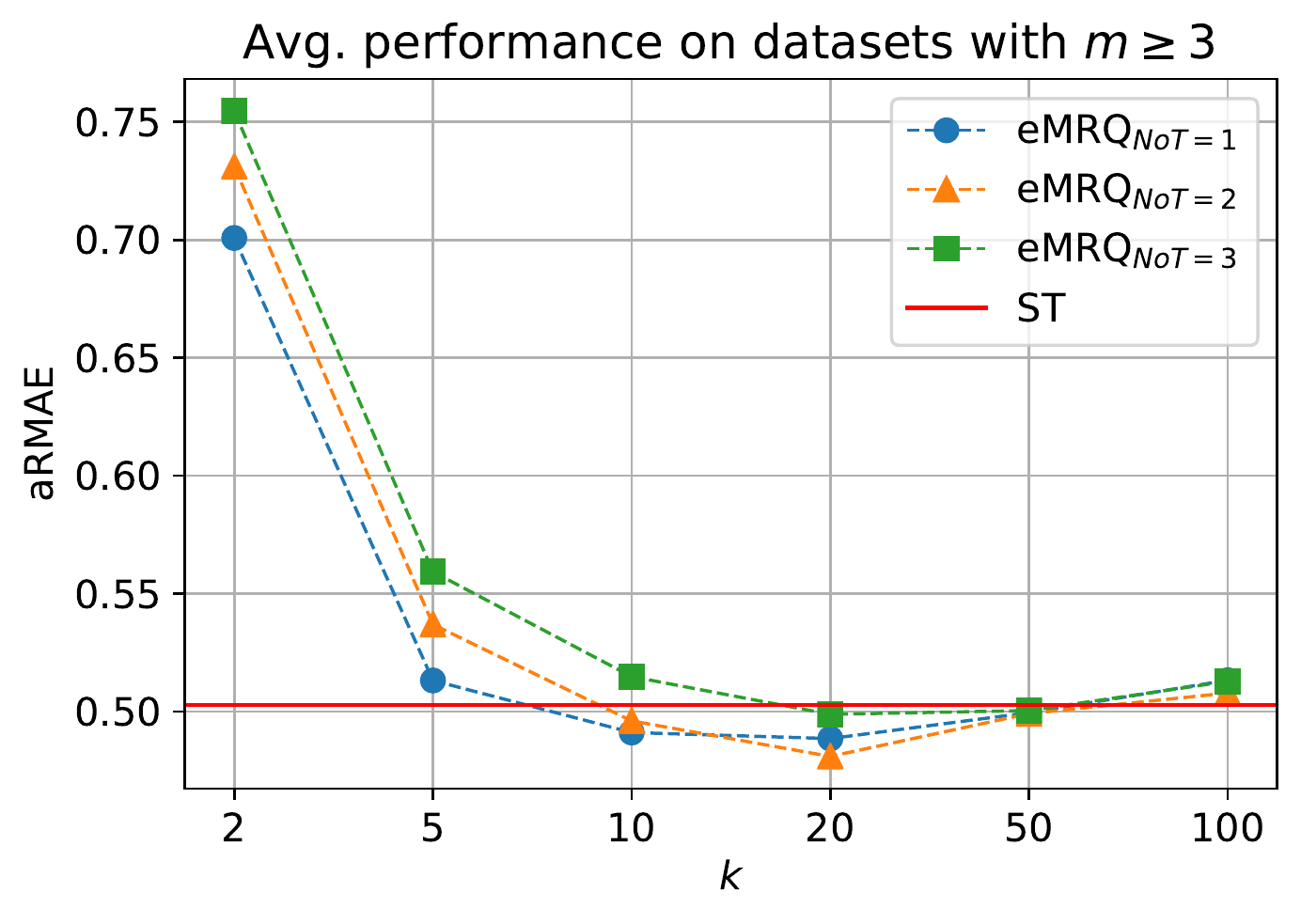}
}
\caption{Performance (aRMAE) of eMRQ as a function of $k$ for $NoT \in \{1,2,3\}$.}
\label{fig:emrq_not}
\end{figure*}

\begin{figure*}[h]
\centering
\resizebox{0.32\textwidth}{!}{
    \includegraphics[trim = 0.0cm 0.1cm 0.1cm 0.2cm, clip]{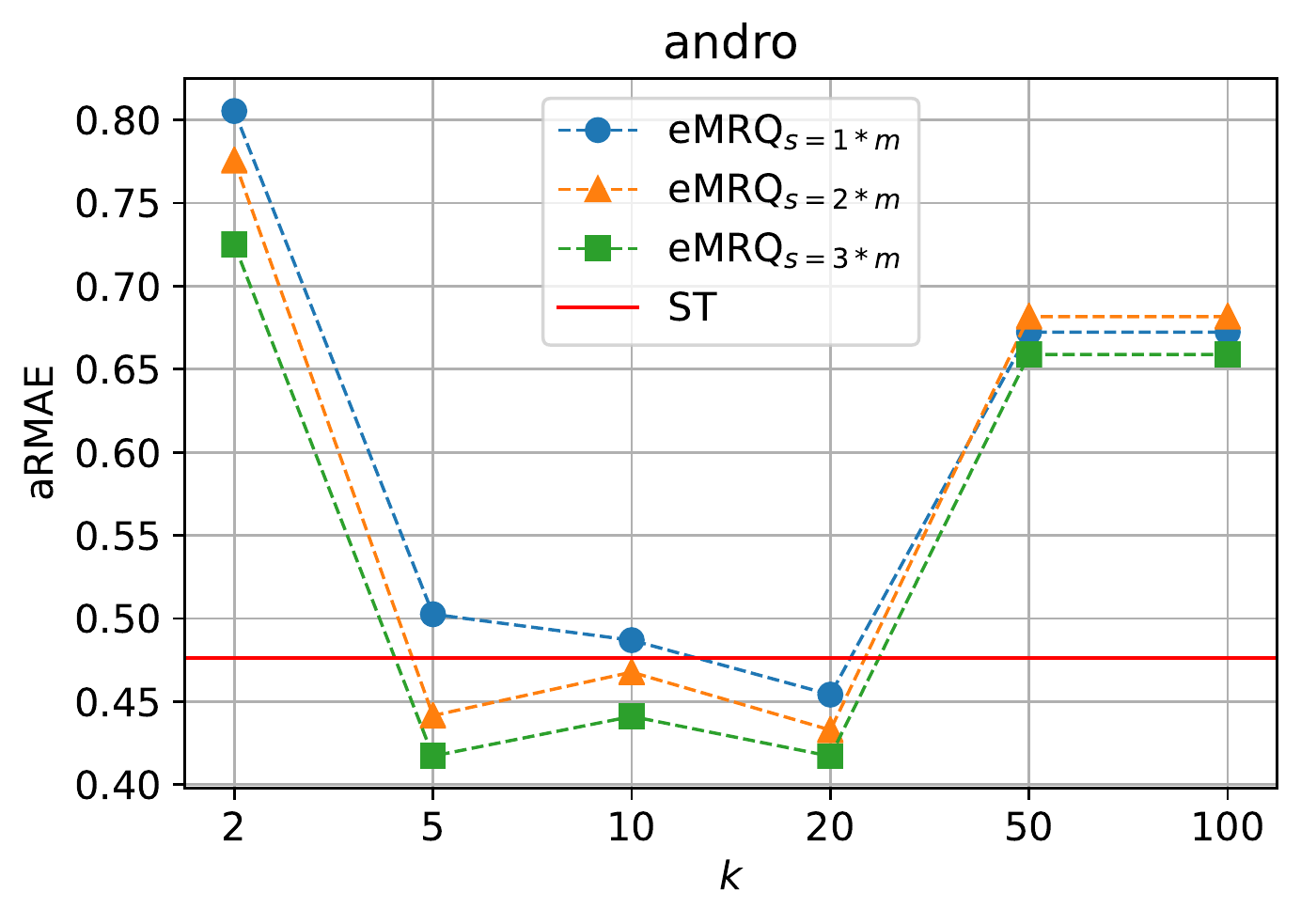}
}
\resizebox{0.32\textwidth}{!}{
    \includegraphics[trim = 0.0cm 0.1cm 0.1cm 0.2cm, clip]{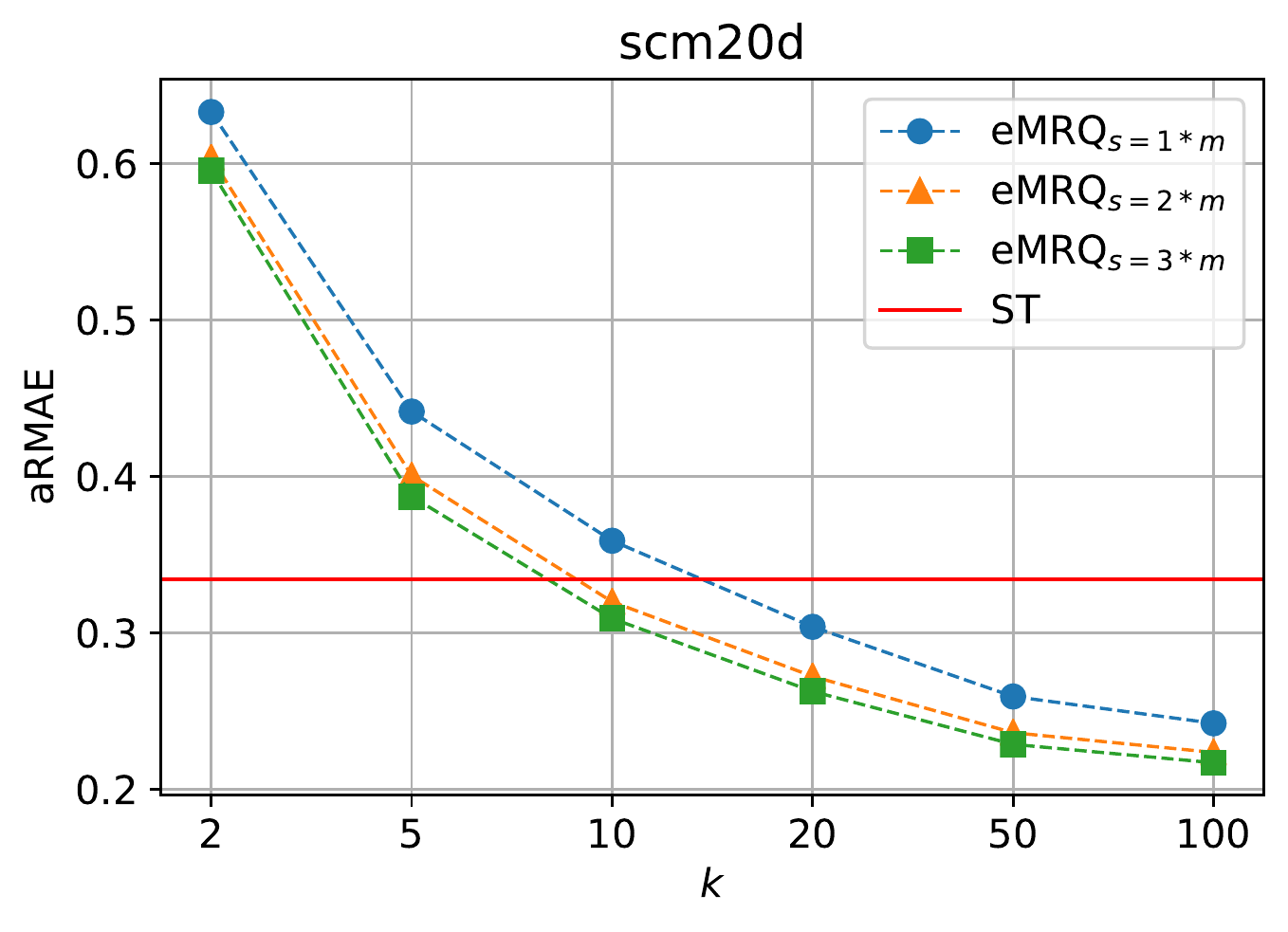}
}
\resizebox{0.32\textwidth}{!}{
    \includegraphics[trim = 0.0cm 0.1cm 0.1cm 0.2cm, clip]{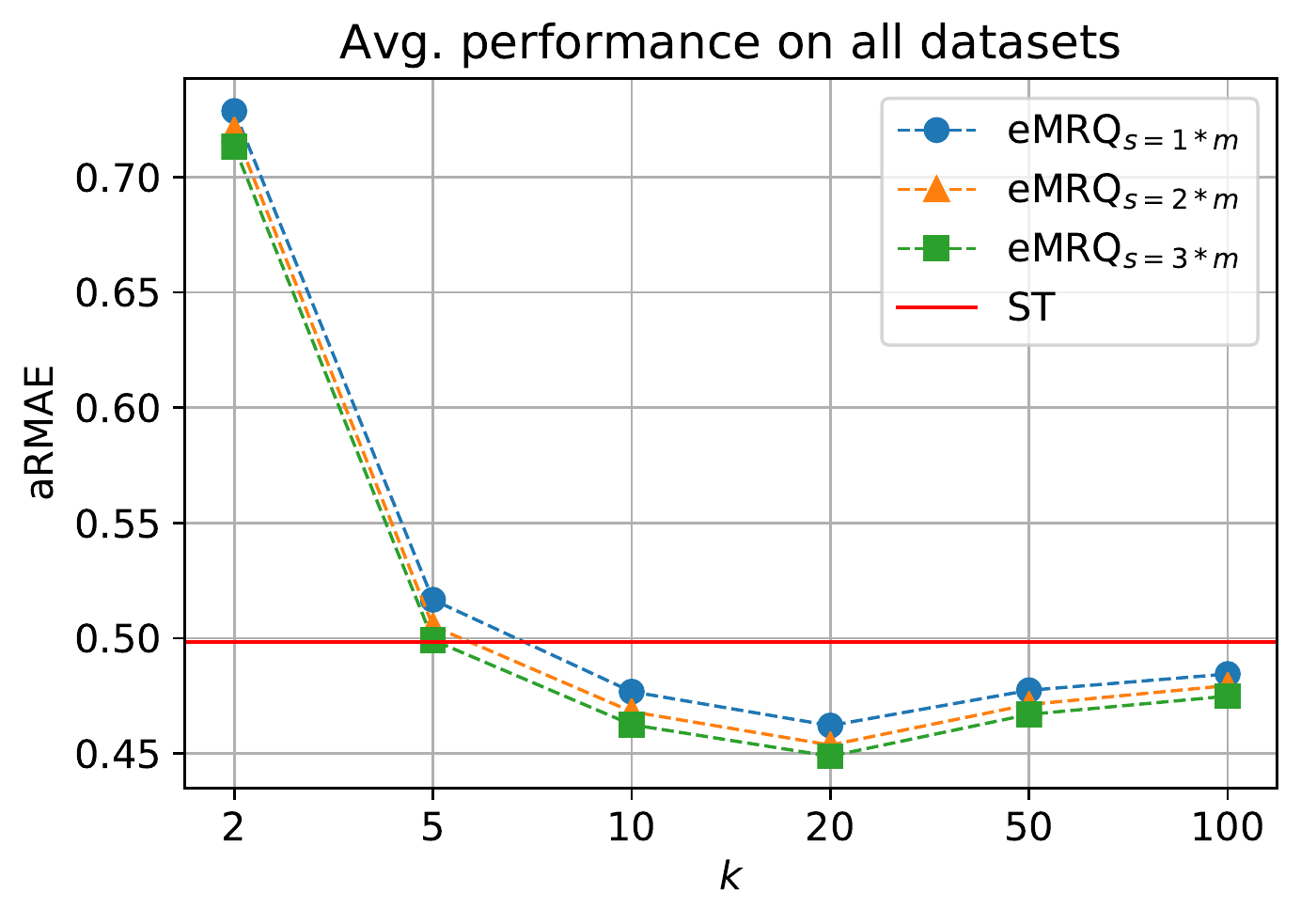}
}
\caption{Performance (aRMAE) of eMRQ as a function of $k$ for $s \in \{m,2m,3m\}$.}
\label{fig:emrq_noq}
\end{figure*}

In this section we study the performance of eMRQ with respect to its parameters: $k$, $NoT$ and $s$. To simplify the analysis, we first fix the $s$ parameter at $s=m$, i.e. we use as many subquantizers as the number of targets in each dataset, and study the interaction between $k$ and $NoT$. Then, we fix the $NoT$ parameter and study the interaction between $k$ and $s$.

Figure~\ref{fig:emrq_not} shows the aRMAE of eMRQ as a function of $k$ for $NoT \in \{1,2,3\}$ on \texttt{andro} and \texttt{scm20d}, as well as the average aRMAE obtained for each combination of values across all datasets (except for \texttt{edm} and \texttt{enb} which have only two target variables and hence $NoT=3$ is not applicable). Note that $NoT=1$ corresponds to independently quantizing each target variable and was included to directly examine the advantages of joint modelling in eMRQ.

Looking at the average performance, we see that joint modelling is indeed advantageous as the best results are obtained with $NoT=2$ and $k=20$. However, larger values of $NoT$ do not lead to better results on average. On the other hand, we also see that a different set of parameters leads to better results in each dataset ($NoT=3$, $k=10$ in \texttt{andro} and $NoT=2$, $k=100$ in \texttt{scm20d}). This suggests that careful per dataset tuning of these parameters is required for optimal performance.  

Figure~\ref{fig:emrq_noq} shows the aRMAE of eMRQ as a function of $k$ for $s \in \{1,2,3\}$ (for $NoT=2$) on \texttt{andro} and \texttt{scm20d}, as well as the average aRMAE obtained for each combination of values across all datasets. We see that the average performance of eMRQ is improving with larger values of $s$ for all values of $k$ and this behavior is consistent across the datasets shown here as well as the remaining datasets. 

\subsection{Comparison with state-of-the-art}
\label{sec:results:soa}

In this section, we compare the performance of MRQ and eMRQ with that of state-of-the-art multi-targer regression methods. 
In particular, the comparison includes ST, RLC \cite{tsoumakas:2014:ecmlpkdd} and ERC \cite{spyromitros2016multi}, using the setup described in section~\ref{sec:experiments:methods}. 

In the case of MRQ, $k=50$ is used based on the analysis of section~\ref{sec:results_mrq}. In the case of eMRQ, we instantiate two variants, one using $s=m$ and one using $s=3m$. However, instead of fixing or tuning the $k$ and $NoT$ parameters, we employ a randomized version of eMRQ (denoted as eMRQr) where for each of the $s$ subquantizers, $k$ is chosen uniformly at random from the range $[50,100]$ and $NoT$ is chosen uniformly at random from the range $[1,2]$. The advantage of this approach is that it avoids the need for parameter tuning - which increases computational time and can be unstable on smaller datasets - and increases the diversity of the eMRQ ensemble. 

\begin{figure}
\centering
\resizebox{0.48\textwidth}{!}{
\includegraphics[trim = 1.5cm 0.5cm 1.2cm 0.2cm, clip]{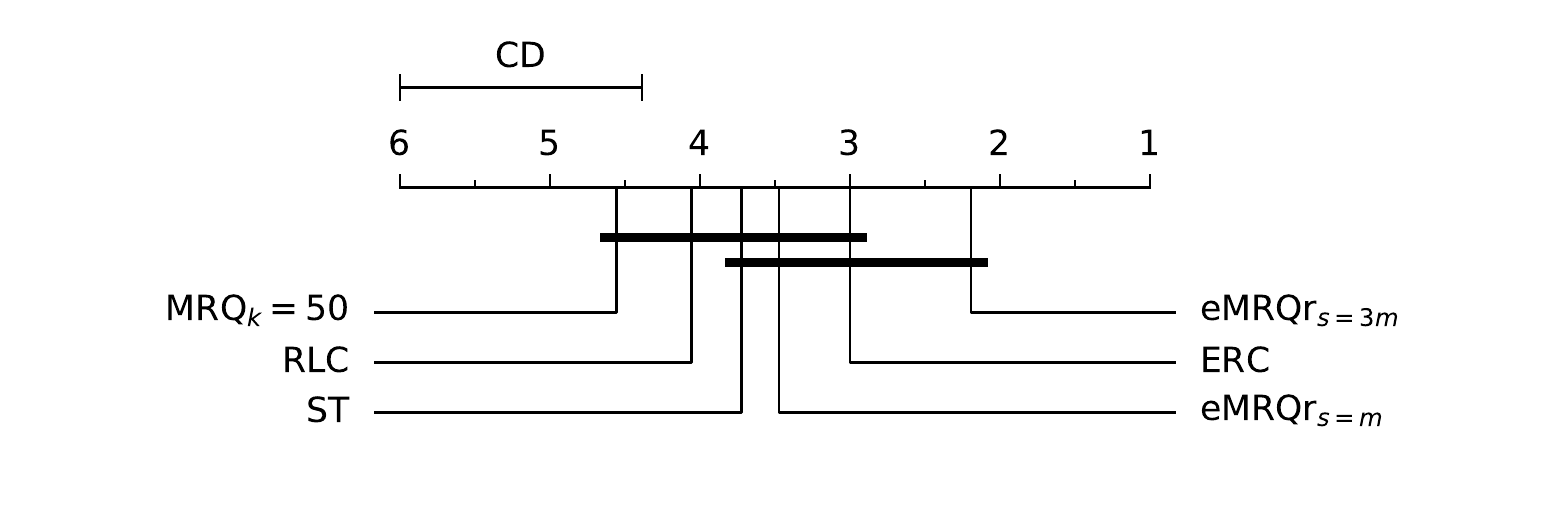}
}
\caption{Critical difference diagram between state-of-the-art methods.}
\label{fig:soa_cd}
\end{figure}

Table~\ref{tbl:soa} shows the results obtained by each method on each dataset, as well as their average ranks and total running times. We first see that MRQ and eMRQr outperform the competing methods in 12 out of 18 datasets. More specifically, MRQ and eMRQr$_m$ have the best performance in two datasets each and eMRQr$_{3m}$ outperforms all other methods in 10 datasets. Looking at the average ranks of the methods, we see that eMRQr$_{3m}$ obtains the lowest average rank, followed by ERC and eMRQr$_{m}$. As can be seen in the critical difference diagram of Figure~\ref{fig:soa_cd}, eMRQr$_{3m}$ performs statistically significantly better than RLC and MRQ, while the experimental data is not sufficient to reach any conclusion with respect to other methods.

The last row of Table~\ref{tbl:soa} reports the total running time\footnote{Experiments were run using 10 cores of a 64-bit CentOS Linux machine equipped with Intel Xeon E7-4860 processors running at 2.27 GHz, leveraging a parallelized implementation of the base learner.} of each method (per dataset running times are omitted due to space limitations). We observe that MRQ is the fastest method overall, while eMRQr$_{3m}$ is an order of magnitude faster than ERC. Taking into account the fact that eMRQr$_{3m}$ is also the best overall performer in terms of accuracy, makes it a very appealing multi-target regression model.  
 
\setlength\tabcolsep{5pt} 

\begin{table}
\caption{Comparison with state-of-the-art. The last two rows show the average ranks of the methods and the total running time in hours respectively.}
\label{tbl:soa}
\begin{center}
\begin{tabular}{l rrrrrr}
\hline
Dataset & ST & RLC & ERC & MRQ$_{50}$ & eMRQr$_{m}$ & eMRQr$_{3m}$ \\
\hline
   edm & 0.822 & 0.817 & 0.823 & \textbf{0.544} & 0.566 & 0.546 \\
   enb & 0.085 & 0.087 & \textbf{0.082} & 0.117 & 0.109 & 0.105 \\
  jura & \textbf{0.529} & 0.546 & 0.531 & 0.658 & 0.664 & 0.563 \\
  scpf & 0.625 & 0.626 & 0.629 & 0.553 & 0.494 & \textbf{0.473} \\
   sf1 & 0.972 & 0.975 & 0.971 & \textbf{0.463} & \textbf{0.463} & \textbf{0.463} \\
   sf2 & 0.883 & 0.927 & 0.868 & 0.462 & \textbf{0.460} & 0.461 \\
 slump & \textbf{0.669} & 0.677 & 0.671 & 0.810 & 0.813 & 0.761 \\
 andro & 0.583 & 0.569 & \textbf{0.548} & 0.870 & 0.747 & 0.726 \\
 atp1d & 0.313 & 0.325 & 0.312 & 0.343 & 0.280 & \textbf{0.261} \\
 atp7d & 0.459 & 0.470 & 0.441 & 0.365 & 0.307 & \textbf{0.301} \\
   rf1 & 0.063 & 0.080 & 0.060 & 0.182 & 0.070 & \textbf{0.051} \\
   rf2 & 0.072 & 0.087 & 0.067 & 0.182 & 0.070 & \textbf{0.050} \\
osales & 0.690 & 0.679 & 0.646 & 0.599 & 0.581 & \textbf{0.567} \\
    wq & 0.857 & 0.851 & 0.858 & 0.873 & 0.702 & \textbf{0.697} \\
 oes10 & 0.412 & \textbf{0.410} & 0.412 & 0.466 & 0.494 & 0.454 \\
 oes97 & 0.563 & 0.566 & \textbf{0.562} & 0.647 & 0.648 & 0.615 \\
 scm1d & 0.271 & 0.267 & 0.252 & 0.412 & 0.266 & \textbf{0.238} \\
scm20d & 0.405 & 0.404 & 0.322 & 0.441 & 0.299 & \textbf{0.266} \\
\hline
Av. rank & 3.722 & 4.056 & 3.000 & 4.556 & 3.472 & \textbf{2.194} \\
Time & 1.27 & 9.89 & 90.86 & \textbf{0.14} & 1.72 & 5.56 \\
\hline
\end{tabular}
\end{center}
\end{table}

\section{Conclusion and future work}\label{sec:conclusion}

We presented MRQ, a new problem transformation approach for multi-target regression that was shown to offer comparable accuracy with state-of-the-art methods, while being significantly more efficient. In addition, a more computationally expensive, ensemble version of MRQ was found to be more accurate than other approaches in most studied datasets. MRQ has two important characteristics that distinguish it from other approaches for multi-target regression: (a) it models a discrete approximation of the joint distribution of the target variables, (b) it is scalable to problems with very large output spaces as it builds a constant number of models. 

In the future, we would like to evaluate MRQ on real-world and synthetic datasets with significantly larger output spaces.
We would also like to perform a deeper theoretical analysis of the two main sources of error in MRQ, i.e the quantization and the classification error, and come up with better ways to choose quantization parameters that strike a good balance between these two error components.
Finally, we would like to explore more sophisticated quantization schemes (e.g. \cite{babenko2014additive,martinez2016revisiting}) that reduce the redundancy between the different subquantizers and are hence able to achieve a smaller quantization error for a fixed quantizer complexity.

\bibliographystyle{IEEEtran}
\bibliography{IEEEabrv,references}

\end{document}